# Unsupervised Textual Grounding: Linking Words to Image Concepts

Raymond A. Yeh, Minh N. Do, Alexander G. Schwing
University of Illinois at Urbana-Champaign
{yeh17, minhdo, aschwing}@illinois.edu

## Abstract

*Textual grounding, i.e., linking words to objects in images, is a challenging but important task for robotics and human-computer interaction. Existing techniques benefit from recent progress in deep learning and generally formulate the task as a supervised learning problem, selecting a bounding box from a set of possible options. To train these deep net based approaches, access to a large-scale datasets is required, however, constructing such a dataset is time-consuming and expensive. Therefore, we develop a completely unsupervised mechanism for textual grounding using hypothesis testing as a mechanism to link words to detected image concepts. We demonstrate our approach on the ReferIt Game dataset and the Flickr30k data, outperforming baselines by $7.98\%$ and $6.96\%$ respectively.*

## 1. Introduction

Textual grounding is an important task for robotics, human-computer interaction, and assistive systems. Increasingly, we interact with computers using language, and it won't be long until we will guide autonomous systems via commands such as 'the coffee mug on the counter' or the 'water bottle next to the sink.' While it is easy for a human to relate the nouns in those phrases to observed real world objects, computers are challenged by the complexity of the commands arising due to object variability and ambiguity in the description and the relations. *E.g.*, the meaning of the term 'next to' differs significantly depending on the context.

To address those challenges, existing textual grounding algorithms [13, 43, 52, 40] benefit significantly from the recent progress in cognitive abilities, in particular from deep net based object detection, object classification and semantic segmentation. More specifically, for textual grounding, deep net based systems extract high-level feature abstractions from hypothesized bounding boxes and textual queries. Both are then compared to assess their compatibility. Importantly, training of textual grounding systems such as [13, 40] crucially relies on the availability of bounding boxes. However, it is rather time-consuming to construct

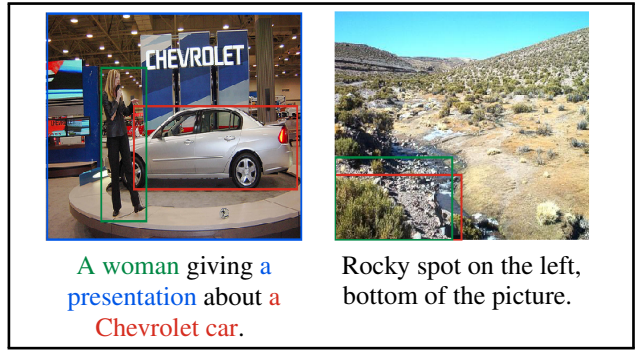

Figure 1. Test set results for grounding of textual phrases from our approach. **Left:** Flickr30k Entities dataset, phrase and bounding box are color coded. **Right:** ReferItGame dataset. (Groundtruth box in green and predicted box in red)

large-scale datasets that facilitate training of deep net based systems.

To address this issue, we propose a completely unsupervised mechanism for textual grounding. Our approach is based on a hypothesis testing formulation which links words to activated image concepts such as semantic segmentations or other spatial maps. More specifically, words are linked to image concepts if observing a word provides a significant signal that an image concept is activated. We establish those links during a learning task, which uses a dataset containing words and images. During inference we extract the linked concepts and use their spatial map to compute a bounding box using the seminal subwindow search algorithm by Lampert *et al.* [26]. Compared to existing techniques, our results are easy to interpret. But more importantly we emphasize that the approach can be easily combined with a supervisory signal.

We demonstrate the effectiveness of the developed technique on the two benchmark datasets for textual grounding, *i.e.*, the ReferIt Game dataset [18] and the Flickr30k data [40]. We show some results in Fig. 1 and we will illustrate that our approach outperforms competing unsupervised textual grounding approaches by a large margin of $7.98\%$ and $6.96\%$ on the ReferIt Game and the Flickr30k dataset respectively.

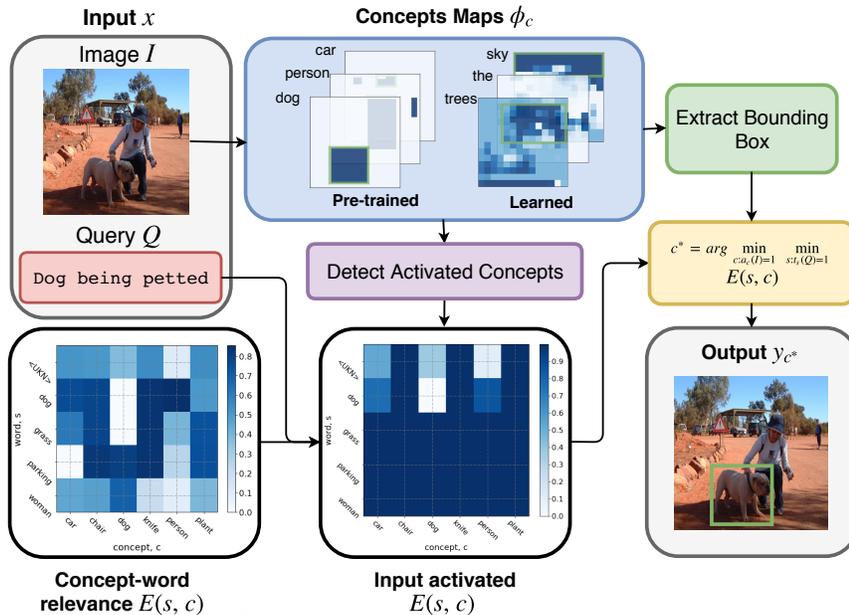

Figure 2. Overview of our proposed approach: We output the bounding box extracted from the active concept that is most relevant to the input query. The relevance of a word-phrase and image concept is learned and represented in $E(s, c)$.

## 2. Related Work

Our method for unsupervised textual grounding combines the efficient subwindow search algorithm of Lampert *et al.* [26] with attention based deep nets. We subsequently discuss related work for textual grounding, attention mechanisms, as well as work on inference with efficient subwindow search.

**Textual grounding:** Textual grounding in its earliest form is related to image retrieval. Classical approaches learn ranking functions via recurrent neural nets [34, 5], metric learning [12], correlation analysis [23], or neural net embeddings [7, 22].

Other techniques explicitly ground natural language in images and videos by jointly learning classifiers and semantic parsers [35, 25]. Gong *et al.* [9] propose a canonical correlation analysis technique to associate images with descriptive sentences using a latent embedding space. In spirit similar is work by Wang *et al.* [52], which learns a structure-preserving embedding for image-sentence retrieval. It can be applied to phrase localization using a ranking framework.

In [10], text is generated for a set of candidate object regions which is subsequently compared to a query. The reverse operation, *i.e.*, generating visual features from query text which is subsequently matched to image regions is discussed in [2].

In [24], 3D cuboids are aligned to a set of 21 nouns relevant to indoor scenes using a Markov random field based technique. A method for grounding of scene graph queries in images is presented in [14]. Grounding of dependency tree relations is discussed in [16] and reformulated using recurrent nets in [15].

Subject-Verb-Object phrases are considered in [45] to develop a visual knowledge extraction system. Their algorithm reasons about the spatial consistency of the configurations of the involved entities.

In [13, 33] caption generation techniques are used to score a set of proposal boxes and returning the hightest ranking one. To avoid application of a text generation pipeline on bounding box proposals, [43] improve the phrase encoding using a long short-term memory (LSTM) [11] based deep net.

Common datasets for visual grounding are the ReferItGame dataset [18] and a newly introduced Flickr30k Entities dataset [40], which provides bounding box annotations for noun phrases of the original Flickr30k dataset [59].

Video datasets, although not directly related to our work in this paper, were used for spatiotemporal language grounding in [28, 60].

In contrast to all of the aforementioned methods which are largely based on region proposal we suggest usage of efficient subwindow search as a suitable inference engine.

**Visual attention:** Over the past few years, single image embeddings extracted from a deep net (*e.g.*, [32, 31, 46]) have been extended to a variety of image attention modules, when considering VQA. For example, a textual long short term memory net (LSTM) may be augmented with a spatial attention [62]. Similarly, Andreas *et al.* [1] employ a language parser together with a series of neural net modules, one of which attends to regions in an image. The language parser suggests which neural net module to use. Stacking of attention units was also investigated by Yang *et al.* [57]. Their stacked attention network predicts the answer successively. Dynamic memory network modules which capture contextual information from neighboring image regions has been considered by Xiong *et al.* [53]. Shih *et al.* [48] use object proposals and and rank regions according to rele-vance. The multi-hop attention scheme of Xu *et al*. [54] was proposed to extract fine-grained details. A joint attention mechanism was discussed by Lu *et al*. [30] and Fukui *et al*. [8] suggest an efficient outer product mechanism to combine visual representation and text representation before applying attention over the combined representation. Additionally, they suggested the use of glimpses. Very recently, Kazemi *et al*. [17] showed a similar approach using concatenation instead of outer product. Importantly, all of these approaches model attention as a single network. The fact that multiple modalities are involved is often not considered explicitly which contrasts the aforementioned approaches from the technique we present.

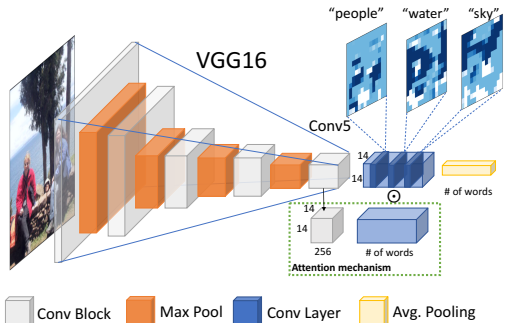

Figure 3. Pipeline of our proposed network for learning image concepts. The network successfully learns the spatial and class information of objects/words in the image and new concepts, *e.g.* classes like "sky" or "water" are not part of the pre-trained classes. Although, the score maps may not be of pixel-accurate quality, extracting useful bounding boxes from them are still feasible.

Very recently Kim *et al*. [20] presented a technique that also interprets attention as a multi-variate probabilistic model, to incorporate structural dependencies into the deep net. Other recent techniques are work by Nam *et al*. [37] on dual attention mechanisms and work by Kim *et al*. [19] on bilinear models. In contrast to the latter two models our approach is easy to extend to any number of data modalities.

**Efficient subwindow search:** Efficient subwindow search was proposed by Lampert *et al*. [26] for object localization. It is based on an extremely effective branch and bound scheme that can be applied to a large class of energy functions. The approach has been applied, among others, to very efficient deformable part models [56], for object class detection [27], for weakly supervised localization [4], indoor scene understanding [47], spatiotemporal object detection proposals [38], and textual grounding [58].

## 3. Approach

We illustrate our approach for unsupervised textual grounding in Fig. 2, where we show how to link a set of 'image concepts,' $c$, (*e.g.*, object detections, and semantic segmentations) to words, $s$, without ever observing any bounding boxes. The image concepts are represented in the form of score maps, which contain both the spatial location, and, when considering the magnitude of the value at every pixel, the strength of the corresponding 'concept' at a particular pixel. By linking a word to an 'image concept,' *i.e.*, by establishing a data-dependent assignment between words and image concepts, we find the visual expression of each word. Importantly, for simplicity of the framework, each word is only assigned to a single concept. The bounding-box accumulating within its interior the highest score of the linked 'image concept' score map is the final prediction.

We refer to capturing the 'image-concept'-word relevance $E(s, c)$ as learning. We propose as a useful cue, statistical hypothesis tests, which assess whether activation of a concept is independent of the word observation. If the probability for a concept activation being independent of a word observation is small, we obtain a strong link between the corresponding 'image concept' and the word.

For inference, given a query and an image as input, we find its linked score map by combining the data statistics $E(s, c)$ obtained during training with image and query statistics. While the query statistics indicate word-occurrences, image statistics are given by 'image concept' activations. To compute the 'image concept' activations we detect a bounding-box on its corresponding score map using a branch-and-bound technique akin to the seminal efficient subwindow search algorithm. If the detected bounding box has confidence greater than 0.5, and covers more than 5% of the image we say that the concept is activated. The confidence of the bounding box is obtained by averaging the probability within the bounding box. We then use the activated words and 'image concepts' to select the activated sub-matrix from $E(s, c)$ which captures statistics about 'image concept'-word relevance. From this submatrix, we select the concept that has the lowest probability of being independent from any of the activated word observations. The bounding box detected on the score map corresponding to the selected concept is the inference result.

Beyond a proper assignment we also need the 'image concepts' themselves. We demonstrate encouraging results with a set of pre-trained concepts, such as object detections, and semantic segmentations. We also obtain a set of 'image concepts' by training a convolutional neural network to predict the probability of a word $s$ given image $I$. By change the architecture's final output layer to spatial average pooling, we obtain a score map for each of the words. We include the score maps if the predicted word accuracy exceeds 50%.

In the following, we first describe the problem formulation by introducing the notation. We then discuss our formulation for learning (*i.e.*, linking words to given 'image concepts') and for computation of the 'image concepts.' Lastly, we describe our inference algorithm (*i.e.*, estimation of the bounding box given a word and image concepts).

### 3.1. Problem Formulation

Let $x$ refer to both the input query $Q$ and the input image $I$, *i.e.*, $x = (Q, I)$. The image $I$ has width $W$ and height $H$. To parameterize a bounding box we use the tuple $y = (y_1, \ldots, y_4)$ which contains the top left corner $(y_1, y_2)$ and the bottom right corner $(y_3, y_4)$. We use $\mathcal{Y}$ to refer to the set of all bounding boxes $y = (y_1, \ldots, y_4) \in \mathcal{Y} = \prod_{i=1}^{4}\{0, \ldots, y_{i,\max}\}$. Hereby $y_{i,\max}$ indicates the maximum coordinate that can be considered for the $i$-th variable ($i \in \{1, \ldots, 4\}$) when processing image $I$.

| Approach | Image Features | Accuracy (%) |
|---|---|---|
| **Supervised** | | |
| SCRC (2016) [13] | VGG-cls | 17.93 |
| GroundeR (2016) [43] | VGG-cls | 26.93 |
| **Unsupervised** | | |
| GroundeR (2016) [43] | VGG-cls | 10.70 |
| GroundeR (2016) [43] | VGG-det | - |
| Entire image | None | 14.62 |
| Largest proposal | None | 14.73 |
| Mutual Info. | VGG-det | 16.00 |
| Ours | VGG-cls | **18.68** |
| Ours | VGG-det | **17.88** |
| Ours | Deeplab-seg | **16.83** |
| Ours | YOLO-det | **17.96** |
| Ours | VGG-cls + VGG-det | **20.10** |
| Ours | VGG-cls + YOLO-det | **20.91** |

Table 1. Phrase localization performance on ReferItGame (accuracy in %).

| Approach | Image Features | Accuracy (%) |
|---|---|---|
| **Supervised** | | |
| CCA (2015) [40] | VGG-cls | 27.42 |
| SCRC (2016) [13] | VGG-cls | 27.80 |
| CCA (2016) [41] | VGG-det | 43.84 |
| GroundeR (2016) [43] | VGG-det | 47.81 |
| **Unsupervised** | | |
| GroundeR (2016) [43] | VGG-cls | 24.66 |
| GroundeR (2016) [43] | VGG-det | 28.94 |
| Entire image | None | 21.99 |
| Largest proposal | None | 24.34 |
| Mutual Info. | VGG-det | 31.19 |
| Ours | VGG-cls | 22.31 |
| Ours | VGG-det | **35.90** |
| Ours | Deeplab-seg | **30.72** |
| Ours | YOLO-det | **36.93** |

Table 2. Phrase localization performance on Flickr 30k Entities (accuracy in %).

The problem of unsupervised textual grounding is the task of predicting the corresponding bounding box $y$ given the input $x$, while the training dataset $\mathcal{D}$ contains only images and corresponding queries, i.e., $\mathcal{D} = \{(x)\}$. We emphasize that no 'bounding box'-query pairs are ever observed, neither during learning nor during inference. Following prior work, pre-trained detection or classification features are assumed to be readily available, but no pre-trained natural language features are employed. We subsequently discuss our formulation for those two tasks, i.e., inference and learning.

### 3.2. Learning

We learn the 'image concept'-word relevance $E(s, c)$ from training data $\mathcal{D} = \{(x)\}$. Importantly, $E(s, c)$ captures the relevance in the form of a distance, i.e., if a word $s$ is related to a concept $c$, then $E(s, c)$ should be small.

We first make some assumptions about the data and the model. We introduce a set of words of interest, $\mathcal{S}$. All other words are captured by a special token which is also part of the set $\mathcal{S}$. We use $t_s(Q) \in \{0, 1\}$ to denote the existence of the token $s \in \mathcal{S}$ in a query $Q$. Additionally, we let $\mathcal{C}$ denote a set of concepts of interest. Further, we use $a_c(I) \in \{0, 1\}$ to denote whether image concept $c$ is activated in image $I$. As mentioned before, we say a concept is activated if the bounding box extracted with efficient sub-window search has a confidence greater than 0.5, and if it covers more than 5% of the image area.

Our 'image concept'-word relevance $E(s, c)$ is inspired by the following intuitive observation: if a word $s$ is relevant to a concept $c$, then the conditional probability of observing such concept given existence of the word, i.e., $P(a_c = 1|t_s = 1)$, should be larger than the unconditional probability $P(a_c = 1)$. For example, let's say the query contains the word "man." We would then expect the probability of the "person" concept to be higher given knowledge that the word "man" was observed. To capture this intuition quantitatively, we perform statistical hypothesis testing for each 'image concept'-word pair.

For each word $s$ and image concept $c$, we construct the following null hypothesis $H_0(c, s)$ and the corresponding alternative hypothesis $H_1(c, s)$:

$$H_0(s, c) : P(a_c = 1|t_s = 1) = P(a_c = 1),$$
$$H_1(s, c) : P(a_c = 1|t_s = 1) > P(a_c = 1).$$

The null hypothesis checks whether the probability of an activated concept is independent of observing an activated word. Note that we don't care about capturing a decrease of $P(a_c = 1|t_s = 1)$ compared to $P(a_c)$. Hence, we perform a one-sided hypothesis test.

Given training data $\mathcal{D} = \{(x)\}$ containing image-query pairs $x = (Q, I)$, we can count how many times the word $s \in \mathcal{S}$ appeared, which we denote $N(s) = \sum_{Q \in \mathcal{D}} t_s(Q)$. Next, we count how many times the word $s$ and the concept $c$ co-occur, which we refer to via $N(s, c) = \sum_{(Q, I) \in \mathcal{D}} a_c(I) \cdot t_s(Q)$. Note that $N(s, c) \leq N(s)$.

We now introduce a random variable $n_{s,c}$ which models the number of times concept $c$ occurs when knowing that sentence token $s$ appears. Assuming $a_c$ to follow a Bernoulli distribution, then $n_{s,c}$ follows the Binomial distribution with $N(s)$ trials and a success probability of $P(a_c = 1|t_s = 1)$, i.e., $P(n_{s,c}) = \text{Bin}(N(s), P(a_c = 1|t_s = 1))$. For large sentence token counts $N(s)$, the Binomial distribution

$$P(n_{s,c}) = \text{Bin}(N(s), P(a_c = 1|t_s = 1)) \approx \mathcal{N}(\mu, \sigma^2)$$

can be approximated by a normal distribution $\mathcal{N}$ with mean $\mu = N(s)P(a_c = 1|t_s = 1)$ and variance $\sigma^2 =$

|  | people | clothing | body parts | animals | vehicles | instruments | scene | other |
|---|---|---|---|---|---|---|---|---|
| # Instances | 5,656 | 2,306 | 523 | 518 | 400 | 162 | 1,619 | 3,374 |
| Entire Image | 27.83 | 5.24 | 0.76 | 17.56 | 25.50 | 15.43 | **45.77** | 16.56 |
| Largest proposal | 31.80 | 7.58 | 2.10 | 30.11 | 34.50 | 17.28 | 41.21 | 17.21 |
| GroundeR, VGG-det (2016) | 44.32 | 9.02 | 0.96 | 46.91 | 46.00 | 19.14 | 28.23 | 16.98 |
| Ours, VGG-det | **61.93** | **16.86** | **2.48** | 64.28 | 54.0 | 9.87 | 16.66 | 14.25 |
| Ours, YOLO-det | 58.37 | 14.87 | 2.29 | **68.91** | **55.00** | **22.22** | 24.87 | **20.77** |

Table 3. Unsupervised phrase localization performance over types on Flickr 30k Entities (accuracy in %).

$N(s)P(a_c = 1|t_s = 1)(1 - P(a_c = 1|t_s = 1))$. Since we use a continuous distribution to approximate a discrete one, we apply the classical continuity correction. Justified by a mean occurrence count of $428$ in our case, we use this approximation for computational simplicity in the following. We note that an exact computation is feasible as well, albeit being computationally slightly more demanding.

To check whether the null hypothesis is reasonable, we assume it to hold, and compute the probability of observing occurrence counts larger than the observed $N(s, c)$. The lower the probability, the more appropriate to reject the null hypothesis and assume $P(a_c = 1|t_s = 1) > P(a_c = 1)$, i.e., we accept the alternative hypothesis $H_1(s, c)$.

Formally, we compute

$$E(s, c) = P(n_{s,c} > N(s, c)|H_0(s, c) \text{ True}), \quad (1)$$

which captures the probability that the value of the random variable $n_{s,c}$ is larger than the observed co-occurence count $N(s, c)$ when assuming the null hypothesis $H_0(s, c)$ to be true. Since we assume $H_0(s, c)$ to be true and because of the approximation of $P(n_{s,c})$ with a normal distribution, we obtain

$$P(n_{s,c} > N(s, c)|H_0(s, c) \text{ True}) = \frac{1}{2} - \frac{1}{2} \operatorname{erf}\left(\frac{N(s, c) - \mu}{\sigma \sqrt{2}}\right),$$

where erf is the error function, mean $\mu = P(a_c = 1)$ and standard deviation $\sigma = \sqrt{N(s)P(a_c = 1)(1 - P(a_c = 1))}$.

Intuitively, if the probability $E(s, c) = P(n_{s,c} > N(s, c)|H_0(s, c) \text{ True})$ is low, we prefer a distribution that is higher at larger counts. Considering the mean of the normal approximation under the assumption $H_0(s, c)$ true, i.e., $\mu = N(s)P(a_c = 1)$, we achieve a higher probability at larger counts if we accept the alternative hypothesis $P(a_c = 1|t_s = 1) > P(a_c = 1)$.

To compute $E(s, c) = P(n_{s,c} > N(s, c)|H_0(s, c) \text{ True})$ we require $P(a_c = 1)$ for which we use the maximum likelihood estimator $N(c)/|\mathcal{D}|$ obtained from the dataset $\mathcal{D}$, while $N(c) = \sum_{(Q,I) \in \mathcal{D}} a_c(I)$.

In summary, our learning formulation retrieves the matrix $E(s, c)$ which captures a signal indicating whether the null hypothesis $H_0(s, c)$, i.e., independence of concept and word, is true. Given the dataset $\mathcal{D}$ we can obtain this matrix for all tokens $s \in \mathcal{S}$ and for all 'image concepts' $c \in \mathcal{C}$.

### 3.3. Inference

Inference is based on a query $Q$, a set of 'image concepts,' $\mathcal{C}$, such as object detections or semantic segmentations, and the matrix $E(s, c)$ which captures the 'image-concept'-word relevance. Each image concept $c \in \mathcal{C}$ is represented by a score map $\phi_c(I) \in \mathbb{R}^{W \times H}$, with image width $W$ and height $H$.

Given these inputs we compute the activated concepts $a_c(I) \in \{0, 1\}$ by first detecting for each score map $c \in \mathcal{C}$ a bounding box via efficient subwindow search and assessing whether it covers at least $5\%$ of the image and whether it has a confidence greater than $0.5$. We obtain the token activations $t_s(Q) \in \{0, 1\}$ directly from the query $Q$.

Based on the activated concepts and the activated word tokens we find that concept $c^*$ which has the lowest evidence of the null hypothesis being true via

$$c^* = \operatorname*{argmin}_{c \in \mathcal{C}: a_c(I)=1} \min_{s \in \mathcal{S}: t_s(Q)=1} E(s, c). \quad (2)$$

We obtain the estimated bounding box estimate for query $Q$ and image $I$ from concept $\phi_{c^*}$ via an efficient subwindow search. It is guaranteed to cover at least $5\%$ of the image while having a confidence larger than $0.5$. We note that, if the lowest $E(s, c)$ is too large, i.e., if there is not enough evidence to reject the null hypothesis, then we will simply return a bounding box corresponding to the entire image.

### 3.4. Image Concepts

It remains to answer which image concepts $\mathcal{C}$ to use. In this section, we describe how to utilize concepts pre-trained from semantic segmentation, object detection, or by directly learning an image classification network from the aforementioned dataset $\mathcal{D}$.

A semantic segmentation framework results in an estimated class probability for each pixel, directly encoding the spatial and concept information. To convert this probability map to a score map, we threshold the probability at 0.5, setting locations greater than 0.5 to 1, otherwise we use -1.

An object detection framework provides a set of boxes with its corresponding class label and estimated probability as output. To use this type of information we first drop all the boxes with confidence less than 0.5. We then create a score map by assigning a value of one to pixel locations within a bounding box, and a value of negative one to the remaining pixels.

An image classification framework provides a probability for each class, and generally doesn't provide any form of spatial information. Therefore, we subsequently describe a modified deep net architecture for classification of words from images, which is more suitable for retaining the spatial information. It can be trained with the information available in the aforementioned dataset $\mathcal{D}$.

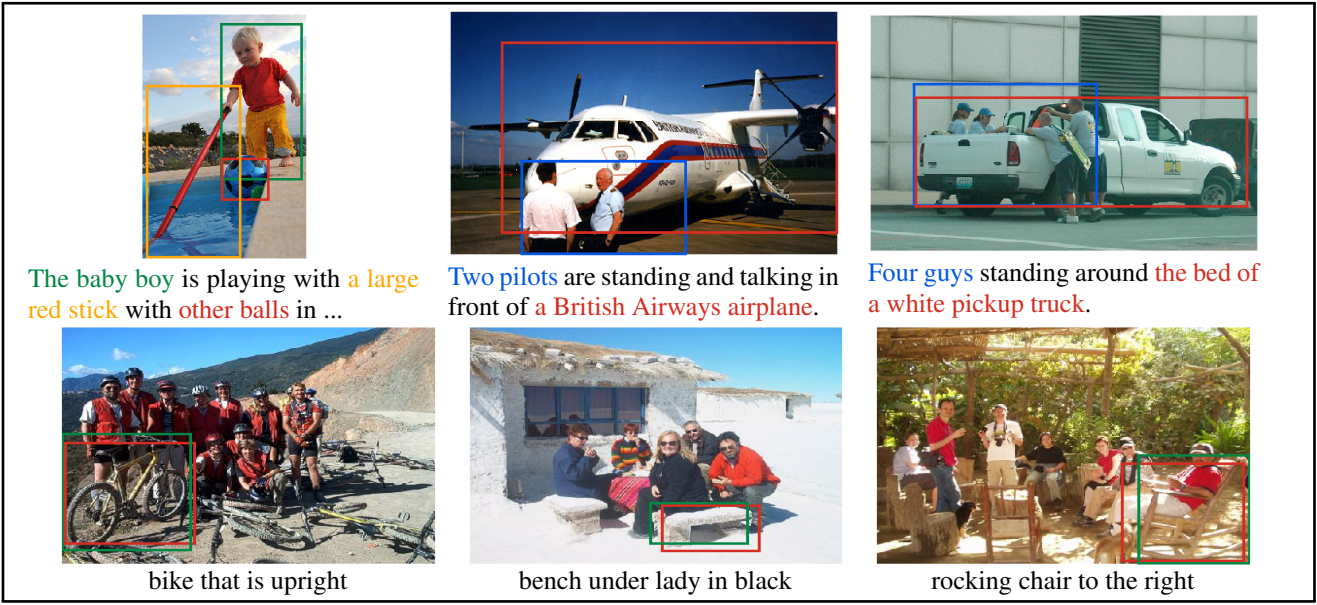

Figure 4. Results from our approach on the test set for grounding of textual phrases. **Top Row:** Flickr 30k Entities Dataset. **Bottom Row:** ReferItGame Dataset (Groundtruth box in green and predicted box in red).

To obtain score maps given an image $I$ and corresponding class labels, *e.g.*, words, we train a deep net with parameters $\theta$, to predict whether a word shows up in the query $Q$ of the corresponding image, *i.e.*, the deep net estimates the probabilities $p_\theta(\hat{t}_s|I)$, $\hat{t}_s \in \{0, 1\}$ for all words $s \in \mathcal{S}$ given an image $I$. The deep net is trained by minimizing the negative cross-entropy between groundtruth $t_s(Q)$ and estimated probability $p_\theta(t_s(Q)|I)$ using the negative binary cross-entropy loss:

$$\mathcal{L} = \sum_{(I,Q)\in\mathcal{D}, s\in\mathcal{S}} t_s(Q) \ln p_\theta(\hat{t}_s|I) + (1 - t_s(Q))(1 - \ln p_\theta(\hat{t}_s|I)). \tag{3}$$

To obtain appropriate score maps $\phi_s(I)$ for each word $s \in \mathcal{S}$, we designed a new output architecture. Our design uses a per channel attention mechanism, *i.e.*, internally, the network outputs a mask in the range of $(0, 1)$ and multiplies it with the hidden activation units. Lastly, we apply global average pooling across the spatial dimension to create the output vector. We extract the activation right before the global average pooling as the score-maps.

In Fig. 3, we visualize the network architecture; we remove the layers of the VGG16 architecture after the Conv5 block and added two convolution layers with 256 channels followed by a convolution layer having $|\mathcal{S}|$ channels to create the attention mask. The attention mask goes through a sigmoid non-linearity to obtain a range of $(0, 1)$. Next, the attention mask is element-wise multiplied with a separate branch of a convolution layer with $|\mathcal{S}|$ number of channels. Lastly, an average pooling across the spatial dimension gives a final vector of dimension $1 \times |\mathcal{S}|$ representing the estimated binary probabilities of each word in $|\mathcal{S}|$.

The main motivation of the architecture is to separately model spatial and semantic information, hence the two branch structure. Unlike classical attention mechanisms, which use a single mask over the image at the input [55], our attention mechanism is 'per class.' Additionally, in order to preserve the spatial information as much as possible, our network architecture is fully convolutional until the very end. A similar architecture has been used for image classification in [50]. Zhou *et al*. [61] also use the average pooling technique on the feature maps to visualize the internal workings of a trained deep network. More advanced network architecture and attention mechanism could be used, but this is beyond the scope of this paper.

Once the network has been trained, we obtain $|S|$ score maps. However, we don't expect all maps to be of acceptable quality. Therefore, we remove the score maps, where the word accuracy on the training set is less than 50%. We provide more visualization of the learned score maps and an ablation study in the experimental section.

## 4. Experimental Evaluation

We first discuss the datasets and provide additional implementation details before discussing the results of our approach.

**Datasets:** The ReferItGame dataset consists of more than 99,000 regions with natural language expressions from 20,000 images. We use the same train, validation, and test split as in [43]. The Flickr30k Entities dataset consists of more than 275k bounding boxes with natural language phrases from 31k images. We use the same train, validation, and test split as in [40].

**Language processing:** To process the free-form textual phrases, we restrict the vocabulary to a fixed set of the most frequent words in the training set and introduce an additional 'unkown' token, `<UKN>`, to represent the remaining words. For the ReferItGame dataset we use the top 200 words and for the Flickr30k Entities we use the top 1000 words. These choices cover ∼90% of all phrases in the training set. We do not consider punctuation, and don't differentiate between lower an upper case.

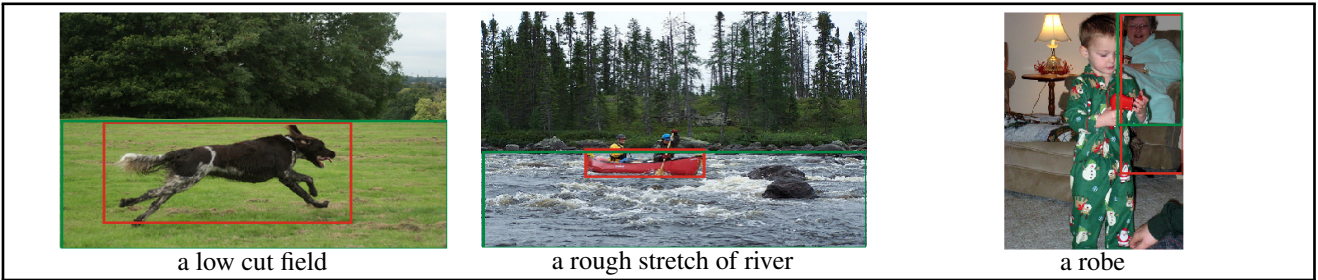

Figure 5. Flickr30k Failure Cases, (Green box: ground-truth, Red box: predicted).

**Image concepts:** Our image concepts can be categorized into two parts, learned concepts and pre-trained concepts.

For the learned concept, we utilize VGG-16 [49] trained on the ILSVRC-2012 classification task [44], which we denote as VGG-cls. We only train the last output layer, and keep the pre-trained parameters fixed. For optimization we use Adam [21] with a learning rate of $1e^{-4}$. We monitor the training and validation log-loss to determine when the model doesn't improve further.

For the pre-traiend concepts, we extract the score maps using a VGG-16 based detection network fine-tuned for object detection on PASCAL VOC-2012 [6], which we refer to as VGG-det. This choice ensures a fair comparison to earlier work.

To illustrate that our approach supports a varierty of features, and to ensure that we are not over-fitting to the score-maps extracted from the VGG architecture, we evaluated on semantic segmentation from the DeepLab system [3] trained on PASCAL VOC-2012 [6], and an additional set of 80 score-maps extracted from the YOLO object detection system [42] trained on MSCOCO [29].

**Quantitative evaluation:** In Tab. 1, Tab. 2 and Tab. 3 we quantitatively evaluate the effectiveness of our approach comparing to state-of-the-art. We use the same accuracy metric as in previous work: prediction is considered to be correct, if the IoU with the ground-truth box is more than 0.5. We observe that our approach, when using the same features, outperforms state-of-the-art unsupervised methods by 6.96% on Flickr30k Entities dataset and by 7.98% on the ReferIt Game dataset.

We also provide an ablation study in Tab. 1 and Tab. 2 showing the effectiveness of each of the components in our system. We observe that object detection features perform better than semantic segmentation based score maps. Bounding boxes extracted from semantic segmentation tend to be too conservative, i.e., the extracted boxes tend to cover only regions within the object and do not include any background, which is not desirable for object bounding boxes. For classification features, on the ReferIt dataset, it learned scenery concepts such as "sky," "grass," and "water," which provided useful spatial information. On the other hand, on the Flickr30k dataset, learned concepts include "man," "people," and "dog." However, we found the score maps for "man" and "people" to also cover other concepts in the background that are highly correlated, e.g., buildings and indoor scenes. Interestingly, the learned "dog" concept is of high fidelity, which is likely due to the large amount of training data on dogs in the ILSVRC-2012 classification dataset. We note that more advanced attention mechanisms and network architectures could be used.

Since our approach boxes the entire image when none of the concepts are relevant, we provided two additional baselines. 'Entire image' denotes the performance, where we always output the bounding box capturing the entire image. 'Largest proposal' refers to picking the largest box out of bounding box proposals. Following [43], we use the top 100 boxes from Selective Search [51] for Flickr30k Entities and the top 100 boxes from Edge Boxes [63] for the ReferIt Game dataset. We want to emphasize that GroundeR [43] was designed as a semi-supervised method and quickly outperforms those baselines once labeled boxes are provided. Inspired by [64], we also compared with another baseline, where $E(s,c)$ is replaced with mutual information.

Next, our approach also generalizes and benefits from having more concepts, i.e., our best system using the score-maps extracted from YOLO attains approximately another 1% improvement.

Lastly, in the unsupervised setting, pre-trained language information is assumed to not be available, i.e. we treat the words and concepts as tokens and do not utilize any word level meaning. Nonetheless we compared with a baseline which uses NLP information by replacing $E(s,c)$ with word vectors extracted from GloVe, pretrained on Wikipedia [39]. Our approach outperforms this baseline by 0.5% on Flickr and 2% on ReferIt with VGG-det features. We suspect that there is a mismatch between word distributions of the grounding datasets and Wikipedia.

**Qualitative evaluation:** In Fig. 4 we evaluate our approach qualitatively, showing the success cases. Our approach successfully captures the objects and scenes described by the query. In Fig. 5, we show failure cases. Our system may fail when concepts are not in the pre-trained or learned concept set. For example, words related to clothing are linked to the person concept, and words for scene are linked to concepts that co-occur with the scene, *e.g.*, river is linked boat, and field is linked to dogs.

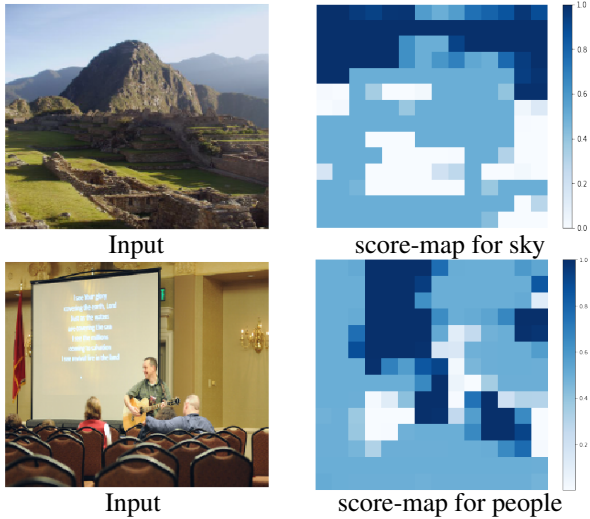

Figure 6. Success and failure case of the learned score-maps. **Top Row:** The learned score map successfully extracts the spatial information of sky. **Bottom Row:** The learned score map fails to pick up the concept of people. The score-map picks up indoor items instead.

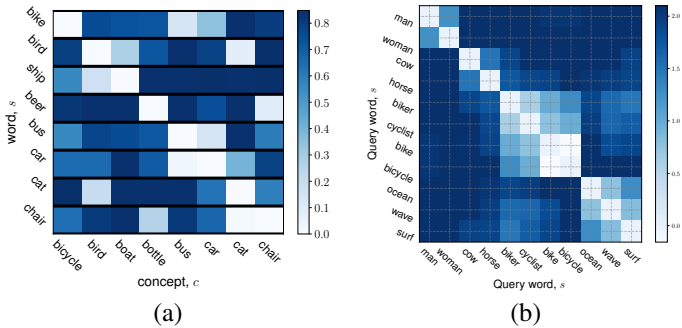

Figure 7. (a) Trained $E(s,c)$ visualized on word, $s$, and detection concepts $c$, (b) Euclidean distance visualized between the vector $E(s,:)$ and $E(s',:)$. Both visualizations are trained on Flickr30k.

| Word  | Most relevant concepts |
|-------|------------------------|
| man   | person, surfboard, toilet, hot dog, tie |
| woman | person, broccoli, handbag, scissors, knife |
| boy   | skateboard, bed, person, sports ball, toothbrush |
| girl  | bed, person, fork, sofa, toothbrush |
| police| bus, motorbike, horse, traffic light, car |
| chef  | knife, spoon, bowl, apple, sink |

Table 4. The top five relevant concepts for each word based on $E(s,c)$ trained on Flickr30k using YOLO-det image features.

In Fig. 6 we show the success and failure case of the learned image concepts. For the ReferIt dataset, the attention mechanism approach successfully captures the spatial scenery concepts (*e.g.*, sky, grass). On the Flickr30k dataset, the approach fails to capture spatial information, the deep network is using indoor information to determine where the word "people" will show up. This is likely caused by the fact that people are typically present in indoor scenes.

**Effectiveness of attention mechanism:** To demonstrate the effectiveness of the attention mechanism discussed in Sec. 3.4, we performed an ablation study; We removed the attention mechanism, and added convolutional layers to match the number of parameters. Without the attention mechanism, the performance drops by 2.11% on the ReferIt Game dataset.

**Interpretability and dataset biases:** Our approach learns a value, $E(s,c)$, capturing the relationship for each word-concept pair. We visualize this quantity in Fig. 7 on a subset of words and concepts. Observe that when $E(s,c)$ is small, the phrase word and image concept are related (*e.g.*, $s$=beer, $c$=bottle). As observed, our approach captures the relevant relationships between phrases and image concepts. Additionally, we can interpret the vector $E(s,:)$ as a word embedding. In Fig. 7 (b) we visualize the Euclidean distance between pairs of word vectors. We clearly observe groups of words, for example (man, woman), (cow, horse), (biker, cyclist), (bike, bicycle), (ocean, wave, surf). In spirit those word embeddings are similar to word2vec [36] where the embedding uses the surrounding words to capture the meaning. In our case, word vectors use the detected objects in an image.

Lastly, $E(s,c)$ can be used to understand dataset biases. In Tab. 4, we list, in order, the most relevant concepts for a subset of the words. For example, we were able to quantify gender biases: man linked to surfboard; woman linked to broccoli; but also other occurrence biases for example: police near some vehicle; chef near knife and spoon.

**Computational Efficiency:** Regarding computational efficiency for inference we note that there are three parts: (1) extracting image features; (2) extracting language features; and (3) computing scores. Our approach requires a single pass for the entire image, whereas GroundeR requires a forward pass for each of the 100 proposal boxes. Each forward pass needs on average 142.85ms. The remaining two parts have negligible contribution to the running time, 21ms for our approach and 1ms for GroundeR.

## 5. Conclusion

The discussed approach for unsupervised textual grounding outperforms competing unsupervised approaches on the two classical datasets by 7.98% (ReferIt Game dataset) and 6.96% (Flickr30K dataset). We think it is important to address this task because labeling of bounding boxes is expensive, and data pairs containing an image and a caption is more readily available.

**Acknowledgments:** This material is based upon work supported in part by the National Science Foundation under Grant No. 1718221, Samsung, and IBM-ILLINOIS Center for Cognitive Computing Systems Research (C3SR). We thank NVIDIA for providing the GPUs for this work.